\pdfoutput=1

\documentclass[conference,letterpaper]{ieeeconf}

\IEEEoverridecommandlockouts                              

\makeatletter
\let\NAT@parse\undefined
\makeatother

\usepackage[numbers,compress]{natbib} 

\usepackage[T1]{fontenc} 

\usepackage[utf8]{inputenc} 

\usepackage{times}
\usepackage{verbatim}
\usepackage{multicol}
\usepackage{multirow}

\usepackage[colorlinks=true, linkcolor=Maroon, urlcolor=Maroon, bookmarks=true]{hyperref}

\usepackage{tabularx}
\usepackage{booktabs}       
\usepackage{microtype}      
\usepackage{color}
\usepackage{colortbl}

\usepackage[usenames,dvipsnames,svgnames,table,x11names,hyperref]{xcolor}

\definecolor{figcapblue}{rgb}{0,0,0.4}
\definecolor{brightgreen}{rgb}{0.4, 1.0, 0.0}
\definecolor{brightlavender}{rgb}{0.75, 0.58, 0.89}
\definecolor{cadmiumred}{rgb}{0.89, 0.0, 0.13}
\definecolor{candyapplered}{rgb}{1.0, 0.03, 0.0}
\definecolor{amethyst}{rgb}{0.6, 0.4, 0.8}
\definecolor{aureolin}{rgb}{0.99, 0.93, 0.0}

\colorlet{mylinkcolor}{blue!100}

\usepackage{url}            

\usepackage{textcase}

\usepackage{graphicx}
\usepackage{epsfig}
\usepackage[tablename=Table,figurename=Figure]{caption}

\usepackage{subcaption}

\usepackage{algorithmicx}
\usepackage{algpseudocode}

\usepackage{amsmath}
\usepackage{amsfonts}
\usepackage{amssymb}
\usepackage{mathrsfs}
\usepackage{nicefrac}       
\usepackage{relsize}
\usepackage{environ}

\DeclareMathSizes{9}{8}{7}{5}
\DeclareMathSizes{10}{8}{7}{5}
\DeclareMathSizes{11}{8}{7}{5}
\DeclareMathSizes{12}{8}{7}{5}

\expandafter\def\expandafter\normalsize\expandafter
{%
    \normalsize
    \setlength\abovedisplayskip{8pt plus 2pt minus 2pt}
    \setlength\belowdisplayskip{8pt plus 2pt minus 2pt}
    \setlength\abovedisplayshortskip{8pt plus 2pt minus 2pt}
    \setlength\belowdisplayshortskip{8pt plus 2pt minus 2pt}
}

\NewEnviron{myequationsmall}{%
    \begin{equation}
    \scalebox{.75}{$\BODY$}
    \end{equation}
    }

\NewEnviron{mymultlinesmall}{%
    \begin{multline}
    \scalebox{.75}{$\BODY$}
    \end{multline}
    }

\makeatletter
\newcommand*{\mycaptionfont}{\@setfontsize\mycaptionfont{8.75pt}{.85em}}
\makeatother

\DeclareCaptionFont{mycaptionfont}{\mycaptionfont}

\definecolor{figcapblue}{rgb}{0,0,0.4}
\newcommand{\cmmnt}[1]{\ignorespaces}

\usepackage{textcomp}

\newcommand{\singlestrquotes}[1]{{\textquotesingle}{#1}{\textquotesingle}}

\begin{document}

\title{Purely Geometric Scene Association and Retrieval \\- A case for macro-scale 3D geometry}

\author{Rahul Sawhney$^{1,2}$, Fuxin Li$^{2}$, Henrik I. Christensen$^{3}$ and Charles L. Isbell$^{1}$
\thanks{$^{1}$ \scriptsize Institute of Robotics \& Intelligent Machines, College of Computing, Georgia Institute of Technology, Atlanta, USA {\tt rsawhney3@gatech.edu, isbell@cc.gatech.edu}~$^{2}$~School of Electrical Engineering and Computer Science, Oregon State University, Corvallis, USA {\tt lif@engr.orst.edu}~$^{3}$~Institute of Contextual Robotics, University of California, San Diego, USA {\tt hichristensen@ucsd.edu}}%
}



%

\maketitle

\begin{abstract}
  We address the problems of measuring geometric similarity between 3D scenes, represented through point clouds or range data frames, and associating them. Our approach leverages macro-scale 3D structural geometry - the relative configuration of arbitrary surfaces and relationships among structures that are potentially far apart. We express such discriminative information in a viewpoint-invariant feature space. These are subsequently encoded in a frame-level signature that can be utilized to measure geometric similarity. Such a characterization is robust to noise, incomplete and partially overlapping data besides viewpoint changes. We show how it can be employed to select a diverse set of data frames which have structurally similar content, and how to validate whether views with similar geometric content are from the same scene. The problem is formulated as one of general purpose retrieval from an unannotated, spatio-temporally unordered database. Empirical analysis indicates that the presented approach thoroughly outperforms baselines on depth / range data. Its depth-only performance is competitive with state-of-the-art approaches with RGB or RGB-D inputs, including ones based on deep learning. Experiments show retrieval performance to hold up well with much sparser databases, which is indicative of the approach's robustness. The approach generalized well - it did not require dataset specific training, and scaled up in our experiments. Finally, we also demonstrate how geometrically diverse selection of views can result in richer 3D reconstructions.

\end{abstract}

\section{Introduction}

The problems of computing similarity and establishing association between range images and/or 3D point clouds of scenes (observed from a viewpoint, henceforth referred to as \textit{scene-views}) is fundamental to robotics and computational perception in general. It plays an important role in a multitude of applications. Loop closure (identifying a place visited earlier in the trajectory) is intrinsic to metric SLAM (\cite{durrant2006simultaneous}). Localizing with respect to a previously reconstructed map or scene model, and relocalizing after a tracking failure (determining sensor pose without pose priors from trajectory) - both are essential for mapping in practice as well. Different forms of the problem are also key to many navigation scenarios, and in perception tasks such as scene-guided search / foraging or location-based context and activity recognition. 

In a minimally restrictive setting, the aforementioned problems (and several others) can be formulated as a retrieval problem - to recognize / identify a scene-view by linking it to stored ones in an assorted, unorganized database. Such a setting would not require any pose priors, spatio-temporal contiguity of collected data\footnote { Spatio-temporally unordered databases can store data acquired from multiple sensors, at multiple times and from disparate locations; could just constitute of snapshots covering scenes of interest.} or other additional information such as annotations or reconstructed models, and would remove the need to learn a specific pose estimator / regressor for each workspace.

While a lot of progress has been made over the years, including in the retrieval domain, competitive scene association approaches in literature have mostly been reliant on (discriminative) appearance information. Relatively few methodologies work well on noisy, imperfect 3D point clouds or depth images from the real world. Often they critically rely on additional pieces of information available in their target scenario - to prune the association hypothesis space, or obtain strong indirect priors on scene similarity, or enable construction of aggregated spatial information structures to allow its estimation (for instance, \cite{labbe2014online, fernandez2013fastPb,  cupec2015place, olson2009recognizingPlaces, choi2015robustIndoorRecon, bosse2013placeRec}) \footnote{~Quite commonly,  approaches rely on spatio-temporal contiguity of frames to to obtain priors or accumulate data structures to ascertain the association.}. Approaches also often operate under limited changes in viewpoint and / or on specific types of scene geometry (such as \cite{steder2011placeRec,fernandez2013fastPb, magnusson2009NDTPlaceRec}) or they solve a simplified 2D problem (such as \cite{himstedt2015geometry2D}). Understandably, methodologies like above are either use case limited or restrictive. Note that approaches like \cite{evangelidis2014generativejointreg} do not ascertain association at all - these directly solve for 3D poses between pre-associated set of data frames.

The dearth of purely 3D geometric scene association approaches in the real world can be primarily attributed to the considerably more ambiguous and challenging depth / range sensing modality. In general, the modality has high local ambiguity and may not be lavish with information on the whole (in contrast to \emph{rgb}). Data acquired from commodity 3D range / depth sensing hardware tends to be particularly noisy as well, has several imperfections. Locally smooth, isomorphic and self-similar nature of typical 3D data from indoor or structural environments makes the problem more difficult. Changes in viewpoint, occlusions and partially overlapping views / content significantly exacerbate the problem further.


\begin{figure*}[t!]
\includegraphics[width=1\textwidth]{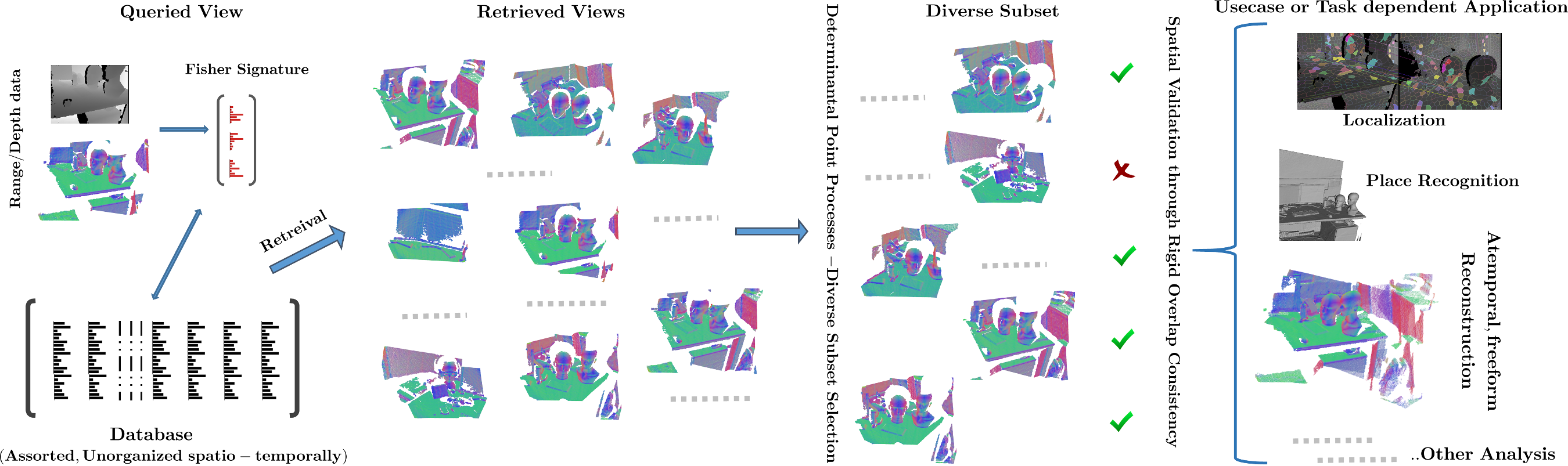}
\caption{\label{fig:pipeline} Retrieval pipeline overview. The query view is indicated in the top-left. Input is a range image or a 3D point cloud. The database (bottom left) constitutes of unordered signatures from arbitrary scene-views, with no labels or ground truth pose annotations. The set of nearest-neighbor retrieved views undergo diversification and subsequent validation. The point clouds are color mapped according to the surface normals - the RGB color of a 3D point is proportional to the component values of its normal. 
}
\vspace*{-1em}
\end{figure*}

We present a minimally restrictive retrieval methodology. Our approach affords means to evaluate \emph{geometric content similarity} between 3D point sets and associate them. We show how it can be utilized to affect \emph{geometric diversity} as well.

We generate descriptive frame-level signatures directly from range images / point clouds (any additional information or assumptions touched upon earlier are not utilized). We make use of macro scale geometry \textemdash ~ 3D geometrical interactions (derived from relative angles and distances) over an arbitrarily large span, between arbitrary surfaces, primitives and structures, and their spatial arrangement (for example between walls, floor and furniture, between fixtures and equipment, or just between various parts of a given structural entity). Such interactions when considered collectively are highly discriminative. They are expressed in a learnt viewpoint invariant feature space (\ref{subsec:feat_representation}). To characterize a scene-view, high order gradient statistics from a dense set of projected interactions are utilized (Fisher Vector, \ref{subsec:encoding}). To identify a geometrically diverse subset from set of similar retrieved views (\ref{subsec:retrieval}), we model a Determinantal Point Process (DPP, \ref{subsec:dpp}). And to establish association with some of the retrieved views, we employ a fine-grained spatial validation scheme which ascertains consistency of rigid geometry overlap (\ref{subsec:validation}).

The proposed approach not only outperformed the range / depth data baseline, but was also comparable or better than state-of-art RGB and RGB-D approaches (including ones based on CNN \footnote{~Note that our approach considers surface patches as far as half a frame apart from the outset -- a distinct difference from popular convolutional network based learning approaches that start by building local features.}) - and without relying on any additional pose annotations, apriori reconstructed 3D world models, or assumptions such as spatio-temporal contiguity of training data used by other approaches.

Experiments also indicated the learning to be general - unlike most other approaches, it did not require dataset specific training; a single learnt model performed well across the board. Experiments also indicated the performance holding up under significantly sparser databases, and under significantly increased database scale and diversity. Our empirical evaluations quantifying geometric diversity of retrievals were quite encouraging as well. They not only indicated a significant increase in viewpoint diversity of the retrieved set, but also suggested the efficacy of the proposed approach for richer reconstruction and increased workspace coverage - promising hitherto unexplored application scenarios, such as assistive structural search.

\section{Related work \label{related_work}}

We refer to only more recent 3D literature among the vast and varied landscape. State-of-the-art loop closure, camera relocalization and place recognition approaches have been primarily based on visual information (\cite{lowry2016visual} presents a recent survey). Many rely on landmark-based features, such as SIFT or ORB, for instance \cite{paul2010fabMap3D, agarwal2011buildingRome, mur2015orbslam, li2015rgbdReloc}. Approaches such as \cite{satkin20133dnn} have focused on the classification problem - one of categorizing similar scenes. \cite{satkin20133dnn} utilizes user annotated 3D data to categorize scenes with viewpoint invariance.


Recent state-of-the-art sensor relocalization approaches in real world structural settings \cite{valentin2015exploiting, brachmann2016dsac, li2015rgbdReloc, walch2017imagelocLSTM, kendall2017geometric} are appearance-reliant as well. They also have other critical requirements like scene-specific learning, and / or workspace models or apriori constructed feature clouds (Section \ref{experiments}).



%

As discussed earlier, high-performing scene association approaches operating solely on 3D range/depth data have been relatively scarce. A significant amount of efforts have been put on local 3D point features, such as \cite{tombari2010SHOT,rusu2009FPFH,steder2011placeRec}. There have also been work based on complete point clouds include variants of Iterative Closest Point, Normal Distributions Transform and aggregated 3D features (often position based, such as height above ground, \cite{granstrom2011learningcloseloops}) over densely sampled keypoints. While they work well under some conditions, their performance deteriorates quickly with increasing change in viewpoint and sensor rotations - \cite{guo2015comprehensive3DdescriptorsIJCV, guo143DfeaturesPAMI, GASP2014_3DV} amongst others, have noted this as well. 


\cite{GASP2014_3DV} matches surface patches between views operating on range / depth data. Our geometric property extraction is along similar lines, and our validation scheme builds upon it. In contrast to \cite{GASP2014_3DV}, which presents a localized surface patch matching algorithm based on aligning geometric sequences defined over neighborhood patches, this work focuses on capturing holistic scene level content for ascertaining geometric content similarity and retrieval.

\begin{figure*}[t!]
\begin{minipage}[h]{0.41\textwidth}
\includegraphics[width=1\textwidth]{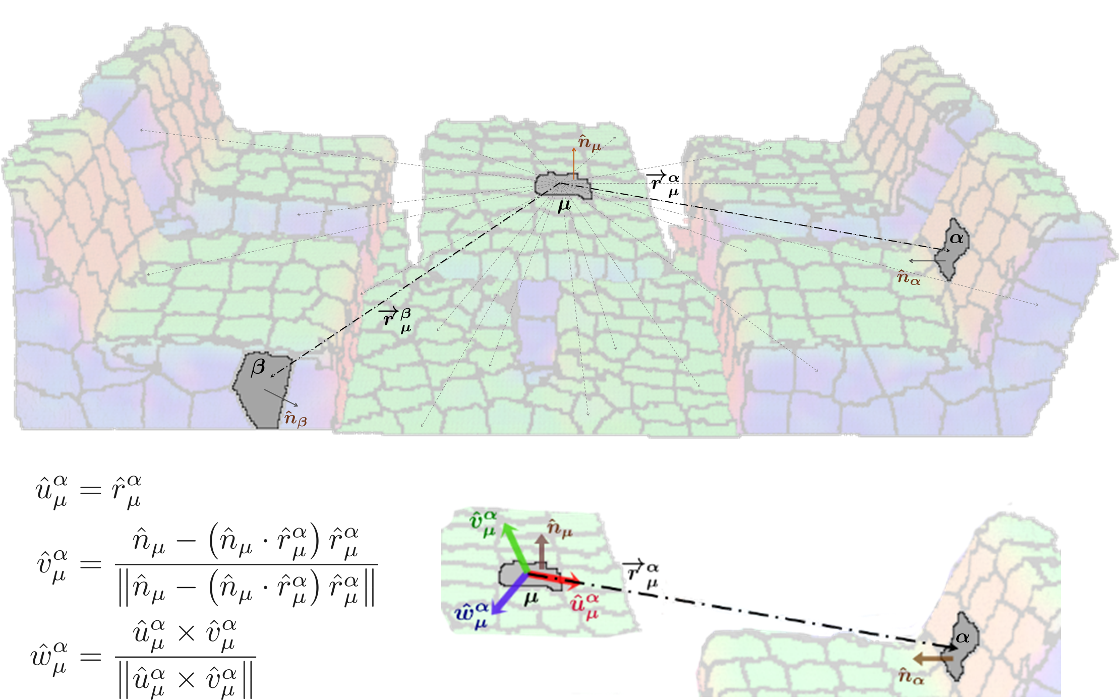}
\caption{\label{fig:feat_extraction} For a given patch $\mu$, relative and invariant 3D properties are extracted with respect to patches in a non-local neighborhood. To facilitate that, an orthonormal, viewpoint agnostic frame is derived using the Gram-Schmidt process.}
\end{minipage}%
\hspace{.4em}
\begin{minipage}[h]{0.59\textwidth}
\includegraphics[width=1\textwidth]{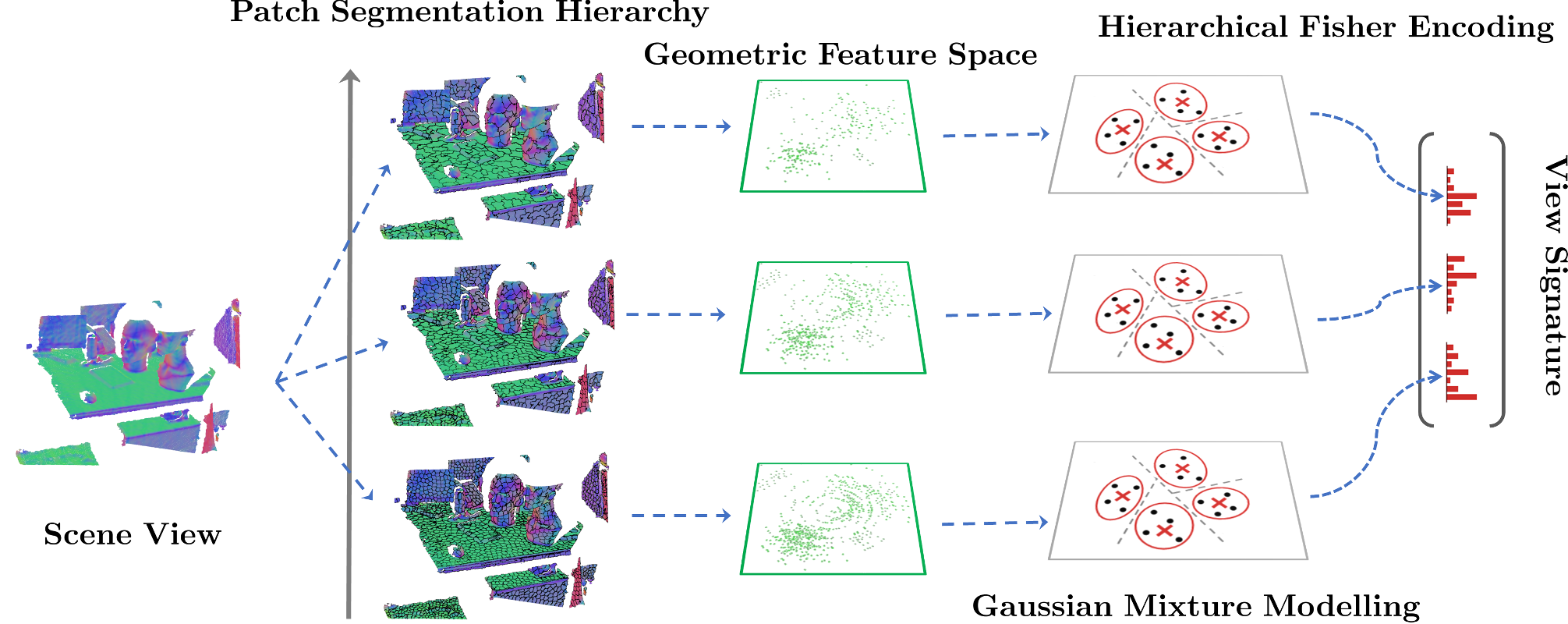}
\caption{\label{fig:scene_encoding}Geometric properties are extracted over a hierarchy of patch segmentations. At each segmentation level, the aggregate sets of properties is first mapped to a viewpoint invariant geometric feature space. These are then jointly encoded as a view level signature using fisher vector embedding.}
\end{minipage}%
\end{figure*}

A number of successful methods exist for shape-based retrieval. \cite{li2015comparison3Dshape} presents a recent survey. Shape retrieval approaches are designed to work with CAD object models or clutter free, object-centric data, often with 3D figure-ground information (in contrast to raw, egocentric scene data from noisy sensors) \footnote{ These also involve specific assumptions - for example, watertight manifolds, surfaces with geometric texture, or disparate / distinctive topology.}. There have been some successful approaches for 3D object instance detection in clutter, by employing pre-ascertained 3D object templates, for example \cite{johnson1999usingSpin, bo2011depthKdesc}. More recently, approaches such as \cite{qi2017pointnet} have learnt object point clouds to identify 3D shapes with distinct topology.

\section{Methodology \label{methodology}}

Given a queried scene-view, $\mathcal{V}_{Q}$, and an extant database, $\mathcal{D}$, of various views from various scenes, $\{\mathcal{V}_{s}\}_{\mathcal{D}}$ - our algorithm ~\textbf{\emph{a)}} Retrieves a set of views which have structurally similar content as $\mathcal{V}_{Q}$,~ \textbf{\emph{b)}} Identifies a geometrically diverse subset of views from this retrieved set, and~ \textbf{\emph{c)}} Ascertains whether some of these views pertain to the same scene as $\mathcal{V}_{Q}$ (Figure~\ref{fig:pipeline}).

We denote $\{x_{i}\}_{i=1}^{c_{X}}$, $x_{i}\in X$ to indicate $\mathcal{V}_{X}$'s segmentation into smooth surface patches. $\left\{ X_{h}\right\} _{h=1}^{H}$ denotes the segmentation hierarchy then. Hierarchy generation is outlined in Section \ref{subsec:details}. To simplify notation, we only indicate the hierarchy level $h$ when it improves clarity.

\subsection{Geometric feature space description \label{subsec:feat_representation}}

\emph{\underline{ Geometric property set extraction}} : For a given view $\mathcal{V}_{X}$, at a particular segmentation level - we first express each patch $x_{i}$ through a $13$-$D$ vector set, $F'_{x_{i}}$ of robust, viewpoint agnostic and macro scale 3D geometric properties. These are derived by utilizing 3D relationships relative to other patches in $x_{i}$'s neighborhood, $\mathcal{N}_{x_{i}}$ (along similar lines as \cite{GASP2014_3DV}). Note that $\mathcal{N}_{x_{i}}$ is large, non-local - it could span the entire segmentation, $X-{x_{i}}$. Neighboring patch count,~$\left|\mathcal{N}_{x_{i}}\right|$, is indicated as $c_{x_{i}}$.
 
 For a patch $x_{i}\equiv\mu\in X$, we denote its mean surface normal as $\hat{n}_{\mu}$ and its 3D mean as  $l_{\mu}$. Denoting $\alpha$ to indicate a patch in $\mu$'s neighborhood, with $\hat{n}_{\alpha}$, $l_{\alpha}$ denoting its normal and mean respectively - an orthonormal basis can be derived from the spanning vectors 
 $\hat{n}_{\mu}$ and $r_{\mu}^{\alpha}=l_{\alpha}-l_{\mu}$ through the Gram-Schmidt process. Figure \ref{fig:feat_extraction} illustrates this. It also formulates the resultant orthonormal basis, $\mathsmaller{<\hat{u}_{\mu}^{\alpha},\,\hat{v}_{\mu}^{\alpha},\,\hat{w}_{\mu}^{\alpha}>}$, where $\hat{u}_{\mu}^{\alpha}$ is the unit vector in the direction of $r_{\mu}^{\alpha}$. Note that coordinate frame spanned by this orthonormal basis is agnostic (invariant) of the sensing viewpoint, since it is a reference frame local to $\mu$ $\mathsmaller{\&}$ $\alpha$. Also note that this basis is seldom degenerate, as $\hat{n}_{\mu}$~and~$r_{\mu}^{\alpha}$ are rarely colinear, especially when data frames are captured from a projective sensing process.
 
 For each neighboring surface patch $\alpha$ in $\mu$'s neighborhood, $\mathcal{N}_{\mu}$, we are able to thus extract the following vector of viewpoint invariant properties, $\mathsmaller{\{f_{\mu}^{'\alpha}\}_{\forall\alpha\in\mathcal{N}_{\mu}}}$ :

\begingroup
\vspace{-.75em}
\small
\begin{multline}
\fontsize{10.25pt}{10.5pt}\selectfont
\hspace{-12pt}f_{\mu}^{'\alpha}=\left[\right.\theta_{\hat{n}_{\alpha},\hat{n}_{\mu}},~\theta_{\hat{u}_{\mu}^\alpha,\hat{n}_{\mu}},~\theta_{\hat{u}_{\mu}^\alpha,\hat{n}_{\alpha}},\,r_{\mu}^{\alpha}\cdot\hat{n}_{\mu},\,\hat{n}_{\alpha}\cdot\hat{u}_{\mu}^{\alpha},\,\hat{n}_{\alpha}\cdot\hat{v}_{\mu}^{\alpha},\,\hat{n}_{\alpha}\cdot\hat{w}_{\mu}^{\alpha},\,\dots
\\r_{\mu}^{\alpha}\cdot(\hat{n}_{\alpha}\times\hat{n}_{\mu}),~||r_{\mu}^{\alpha}||, ||r_{\mu}^{\alpha}||\cdot sgn_{\epsilon_{\theta}}(\hat{n}_{\mu}\cdot\hat{u}_{\mu}^{\alpha}),~||r_{\mu}^{\alpha}||\cdot sgn_{\epsilon_{\theta}}(\hat{n}_{\alpha}\cdot\hat{u}_{\mu}^{\alpha}),\,\dots
\\\hspace{-14pt}||r_{\mu}^{\alpha}||\cdot sgn_{\epsilon_{\theta}}(\hat{n}_{\alpha}\cdot\hat{v}_{\mu}^{\alpha}),\,||r_{\mu}^{\alpha}||\cdot sgn_{\epsilon_{\theta}}(\hat{n}_{\alpha}\cdot\hat{w}_{\mu}^{\alpha}),\,\left.\right]^{T}
\label{eq:coord}
\end{multline}
\endgroup

The $\theta$ above refers to the angle between the indicated vectors and $\times$ represents an outer product. $sgn_{e}(..)$ is a robust signum function that clamps to zero when its parameter $\not\in[cos^{-1}(PI-e_{\theta}),\,cos^{-1}(e_{\theta})]$, with $e_{\theta}$ accounting for allowable tolerance to angular noise.

The feature vector $f_{\mu}^{'\alpha}$ basically represents an overcomplete characterization of relative properties between the two patches - formulated in a viewpoint agnostic fashion. The first part (first $9$ features) captures angular relationships between $r_{\alpha}^{\mu}$, $\hat{n}_{\alpha}$ $\mathsmaller{\&}$  $\hat{n}_{\mu}$,~characterizes $\hat{n}_{\alpha}$ in the invariant frame derived from $r_{\mu}^{\alpha}$~and~$\hat{n}_{\mu}$, and characterizes $r_{\alpha}^{\mu}$. The second part (remaining $4$) consists of robustified features - as a measure against noises arising due to estimation from real world, noisy data. Signs of projected normals' components are captured through robust signum functions and are augmented with the magnitude of relative displacement vector.



\emph{\underline{Feature space projection}} : A patch's property set $\mathsmaller{F_{x_{i}}^{'}=}$ $\mathsmaller{\{f_{x_{i}}^{'\alpha}\}_{\alpha\in\mathcal{N}_{x_{i}}}}$ is then projected onto a subspace learnt through nonlinear independent component analysis (\cite{koldovsky2006efficientICA}). The projection reduces redundancy in $f_{x_{i}}^{'\alpha}$, denoising and making the components more independent. Importantly, this fits with the component independence assumption made in Section~\ref{subsec:encoding} to train Gaussian mixtures with diagonal covariances.

The feature space projection results in a $12$-$D$ feature vector set $\mathsmaller{F_{x_{i}}=\{f_{x_{i}}^{\alpha}\}_{\alpha\in\mathcal{N}_{x_{i}}}}$. By considering the patches in ${x_{i}}$'s macro scale neighborhood, $\mathcal{N}_{x_{i}}$, the  feature set $F_{x_{i}}$ can thus robustly express the 3D geometry in ${x_{i}}$\textquoteright s non-local neighborhood. An aggregation of such feature sets arising from all the patches,~$\mathsmaller{F^{X}=}$ $\mathsmaller{\{F_{x_{i}}\}_{i=1}^{c_{X}}\equiv}$ $\mathsmaller{\{f_{x_{i}}^{\alpha}|x_{i}\in X,\alpha\in\mathcal{N}_{x_{i}}\}}$, can thus invariantly and richly express the geometry of the entire scene as captured by $\mathcal{V}_{X}$. Finally, the above procedure is repeated for each level in the segmentation hierarchy, to capture fine as well as coarse details. This results in a hierarchy of aggregate feature vector sets, $\mathsmaller{\left\{ F_{h}^{X}\right\} _{h=1}^{H}}$.


\subsection{Encoding feature space statistics\label{subsec:encoding}}

To obtain a descriptive signature for a given view, $\mathcal{V}_{X}$, we encode the aggregated feature sets using Fisher vector embedding (FV, \cite{jaakkola1999exploiting,perronnin2010improvingFK}) - this captures the normalized gradient of the log-likelihood of the feature sets. The Fisher kernel theory, first presented in \cite{jaakkola1999exploiting}, introduces a similarity kernel, arising as a consequence of maximizing the log-likelihood of generatively modeled data. In this paper, Gaussian Mixture Models (GMM) were used to model the feature space distribution.

Given a learnt GMM, $P_{\Theta}$, parameterized as $\mathsmaller{\Theta=}$ $\mathsmaller{\left\{ p_{g},\,\nu_{g},\,\Lambda_{g}\right\} _{1}^{G}}$, the FV embedding of the aggregate feature set $F^{X}$, indicated as $\phi(F^{X})$, is obtained as $\mathsmaller{\phi(F^{X})=L_{\Theta}\nabla_{\Theta}log(P_{\Theta}(F^{X}))}$. Here, $L_{\Theta}$ is the Cholesky decomposition factor of the inverse Fisher Information Matrix, and $\mathsmaller{\nabla_{\Theta}log(P_{\Theta}(F^{X}))}$ is the score function (log-likelihood gradient). Following similar analysis as \cite{perronnin2010improvingFK}, under assumptions of diagonal covariance matrices, $\Lambda_{g}$, and independence of the samples, $f_{x_{i}}^{\alpha}$, the embedding evaluates as $\mathsmaller{\phi(F^{X})=}$ $\mathsmaller{\left[m_{1}^{0},\,m_{1}^{1^{T}},\,m_{1}^{2^{T}}\dots m_{g}^{0},\,m_{g}^{1^{T}},\,m_{g}^{2^{T}}\dots m_{G}^{0},\,m_{G}^{1^{T}},\,m_{G}^{2^{T}}\right]^{T}}$ -  where $m_{g}^{0},\,m_{g}^{1},\,m_{g}^{2}$ respectively capture the normalized zeroth, first and second order statistics of the sample set that falls in the $g$-th mixture component of the GMM.  $\phi(F^{X})$ has a dimensionality of $d_{\phi}=(2d_{F}+1)\cdot G$, where $G$ is number of mixture components, and $\emph{\ensuremath{d_{F}=12}}$ is the dimensionality of our geometric feature space. Below, $\mathbf{1}$ denotes an all-one vector and $\mathsmaller{\pi_{ij,g}=\frac{\exp\left[-\frac{1}{2}(f_{x_{i}}^{x_{j}}-\nu_{g})^{T}\Lambda_{g}^{-1}(f_{x_{i}}^{x_{j}}-\nu_{g})\right]}{\sum_{g=1}^{G}\exp\left[-\frac{1}{2}(f_{x_{i}}^{x_{j}}-\nu_{g})^{T}\Lambda_{g}^{-1}(f_{x_{i}}^{x_{j}}-\nu_{g})\right]}}$.

\begingroup
\begin{subequations} 
\footnotesize
\begin{flalign} 
m_{g}^{0}=&\frac{1}{c_{X}c_{x_{i}}\sqrt{p_{g}}}\sum_{i=1}^{c_{X}}\sum_{j=1}^{c_{x_{i}}}(\pi_{ij,g}-p_{g})\\ m_{g}^{1}=&\frac{1}{c_{X}c_{x_{i}}\sqrt{p_{g}}}\sum_{i=1}^{c_{X}}\sum_{j=1}^{c_{x_{i}}}\pi_{ij,g}\Lambda^{-\nicefrac{1}{2}}(f_{x_{i}}^{x_{j}}-\nu_{g})\\ m_{g}^{2}=&\frac{1}{c_{X}c_{x_{i}}\sqrt{2p_{g}}}\sum_{i=1}^{c_{X}}\sum_{j=1}^{c_{x_{i}}}\pi_{ij,g}\left[\Lambda^{-1}(f_{x_{i}}^{x_{j}}-\nu_{g})(f_{x_{i}}^{x_{j}}-\nu_{g}){}^{T}-\mathbf{{I}}\right]\mathbf{1} 
\end{flalign} \label{eq:fisherorderstats} 
\vspace{-11pt} 
\end{subequations}
\endgroup

$\phi(F^{X})$ is then component-wise square root normalized (by replacing each component, \singlestrquotes{$a$} of $\mathsmaller{\phi(F^{X})}$ by \singlestrquotes{$|a|^{\nicefrac{1}{2}}sign(a)$}), and $\ell_{2}$ normalized. The square root normalization serves to alleviate the dominant effect of relatively indiscriminate samples occurring with high frequency (for example, arising from patches on a wall or ceiling) and the $\ell_{2}$ normalization helps generalization across different scenes by normalizing the energy content. The desired view signature vector for $\mathcal{V}_{X}$, denoted as $\psi(X)$, is obtained by evaluating the embedding at each level in hierarchy, and concatenating them --- $\mathsmaller{\psi(X)=}$ $\mathsmaller{\left[\phi(F_{1}^{X})^{T},\dots\,\phi(F_{h}^{X})^{T},\,\dots,\,\phi(F_{H}^{X})^{T}\right]^{T}}$

\subsection{Similarity and Retrieval\label{subsec:retrieval}}

The thus obtained view signature, $\psi(X)$, captures discriminative 3D geometrical properties, and is robust to  viewpoint changes, sensor noise, occlusions and other data imperfections by design. As experiments indicate, a metric based on such view signatures is a reliable measure of 3D geometric similarity. We tried ${\ell}_1$ \& ${\ell}_2$ distance metrics, and used ${\ell}_1$ for all experiments in the paper as it performed better. Thus the similarity between two given views $V_X$ \& $V_Y$ can be denoted as,  $\mathbf{s}(X,Y)=-(\sum_{1}^{25GH}\left|\psi(X)-\psi(Y)\right|_{1}).$ 

Given a a queried view, $\mathcal{V}_{Q}$, and a database $\mathcal{D}$ of view signatures, one can thus retrieve a set of putative view associations in the geometric sense through nearest neighbor
queries. We indicate this retrieved set of putatively associated views as $\mathcal{R}=\{\mathcal{V}_{X}\}_{X=1}^{c_{\mathcal{{R}}}}$.


\subsection{Diversity Sampling with Determinantal Point Processes\label{subsec:dpp}}

Depending on the distribution of scenes' views in the database, $\mathcal{R}=\{\mathcal{V}_{X}\}_{X=1}^{c_{\mathcal{{R}}}}$
could be overwhelmed with views which are \emph{near duplicates} (all being very similar to each other, hence almost equally similar to the queried view). This may not be desirable since the subset of top
retrievals could just be flooded with near duplicates of false putative associations, resulting in complete failure. By filtering out near duplicates, a diversity based subset selection procedure may still be able to salvage correct, albeit lower ranked, putative associations present in $\mathcal{R}$ with further post-processing validation.
 
A diverse set of retrievals is generally desirable. It would provide assorted and possibly complementary
information, which could be made use of thereon. For instance, it could be potentially beneficial in reconstruction or coverage tasks, where diverse viewpoints observing the environment with only partially overlapping content are more desirable than having redundant views from nearly the same perspective. 
A querying human user could also be better assisted by being provided with a diverse set of the retrievals to choose from.

Determinantal point processes (\cite{kulesza2012dppTUT}) are employed to select a diverse subset of candidate views, $\mathcal{{C}}$, from $\mathcal{R}$. A point process~$\mathcal{{P}}_{L}$~is called an $L$ - ensemble $k$-\emph{determinantal point process} if for every random subset, $\mathcal{C}$, of $\mathcal{{R}}$, such that $|\mathcal{{C}}|=k$, drawn according to $\mathcal{{P}}_{L}$, we have $ \mathcal{{P}}_{L}\left(\mathcal{{C}};\mathcal{{R}}\right)=\frac{det(L_{\mathcal{{C}}})}{\Sigma_{\forall\mathcal{{A}}\in\mathcal{{R}},\,|\mathcal{{A}}|=k}\,det(L_{\mathcal{{A}}})}$. $L$ here is a symmetric positive semi-definite similarity matrix indexed
by the elements of $\mathcal{R}$. $L_{\mathcal{{C}}}$ is the principal
minor (submatrix) with rows and columns from $L$ indexed by the elements
in subset $\mathcal{C}$. Thus the probability of selecting a subset
$\mathcal{C}$, ($|\mathcal{{C}}|=k=c_{\mathcal{{C}}}$) elements is directly proportional to the determinant of the submatrix indexed by it. Note that higher diagonal values would proportionately encourage their inclusion in a selected subset $\mathcal{C}$ as they lead to higher determinants. Similarly, the off-diagonal values determine
correlation between different elements, and a high value decreases the determinant overall. Thus two elements with a high similarity value tend not to co-occur in $\mathcal{C}$. DPP sample sets are therefore able to balance the net significance of their constituent elements with their diversity. We modeled $L$ accordingly as follows \vspace{-1.5em}

\begin{equation}
\{L\}_{X,Y}=\rho_{X}\rho_{Y}\kappa e^{\frac{\mathbf{s}(X,Y)}{\sigma}}\,\,,\,\,1\leq X,Y\leq c_{\mathcal{{R}}}
\label{eq:dppmodel} 
\end{equation}

where $\rho_{X}=e^{\frac{1}{2}\frac{\mathbf{s}(X,Q)}{\omega}},\,\exists X\in\mathcal{{R}}$
models the similarity of a retrieved view $\mathcal{V}_{X}$ to the queried view $\mathcal{V}_{Q}$. The similarity between two given
views $\mathcal{V}_{X}$ and $\mathcal{V}_{Y}$~is captured by the rightmost term. Positive valued parameters $\sigma$, $\omega$ and
$\kappa$ can be tuned to balance the need for both diversity and similarity to $\mathcal{V}_{Q}$. A lower sigma would induce a higher resolution in similarity scores between retrieved views, and hence would result in a more diverse subset selection.%


While the \emph{MAP }inference on $\mathcal{{P}}_{L}$~to determine the most probable subset is NP-hard, efficient sampling algorithms exist which provide good approximate solutions in practice. For our purposes, a greedy procedure based on \cite{kulesza2012dppTUT} which results in ${\mathcal{O}(k\log{k})}$-approximation worked well.

\begin{figure}[t!]
\includegraphics[width=1\linewidth]{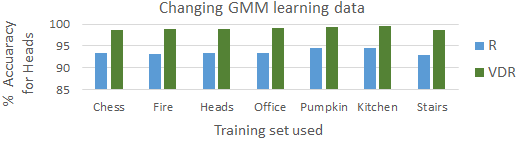}\\[-14pt]
\caption{ Consistency in GMM learning. Similar retrieval accuracies were achieved with GMMs learnt from each of the $7$ training sets.\cmmnt{~ \emph{Bottom} : The impact of encoding a fine to coarse hierarchy of levels. As can be seen, significant improvements are achieved when properties are captured at multiple scales.}} \label{fig:ablations}
\vspace*{-1em}
\end{figure}

\subsection{Validating candidate views for association\label{subsec:validation}}

We employ a finer grained spatial validation step before finally associating the queried view with some of the views in the candidate set, $\mathcal{C}=\{\mathcal{V}_{X}\}_{X=1}^{c\mathcal{_{{C}}}}$ . This is done by directly leveraging the rigid 3D spatial arrangement
of surface patches to ascertain surface alignment. We make use of the patch matching scheme presented in our prior work \cite{GASP2014_3DV}. It utilizes a sequence alignment scheme over similarly motivated patch properties to find standalone correspondences based on 3D neighborhood similarity. A semi-dense set of correspondences can be ascertained. Rigid transform between two views of a given scene can then be robustly, accurately computed through consensus of patch associations. 

When views from scenes with different geometrical content are matched through \cite{GASP2014_3DV}, the matches would likely be inconsistent with respect to the computed transform. We exploit this understanding to validate associations with candidate views. For each candidate view, $\mathcal{V}_{X}\in\mathcal{{C}}$, and the queried view, $\mathcal{V}_{Q}$, we utilize randomly sampled patches to estimate rigid transformations both ways, that is,~$\mathsmaller{T_{Q}^{X}\equiv(R_{Q}^{X},\,t_{Q}^{X})}$ and $\mathsmaller{T_{X}^{Q}\equiv(R_{X}^{Q},\,t_{X}^{Q})}$ and check whether they are consistent with each other. We ascertain a candidate $\mathcal{V}_{X}\in\mathcal{C}$ as associated with $\mathcal{V}_{Q}$ when $\mathsmaller{\left\Vert \log\left(R_{Q}^{X}R_{X}^{Q}\right)\right\Vert _{2}\leq\epsilon_{val}^{\theta}\,\,\&\,\,\left\Vert t_{Q}^{X}+t_{X}^{Q}\right\Vert _{2}\leq\epsilon_{val}^{\varepsilon}}$ - we are basically ensuring that the magnitude of the rotation and translation components in the residual transform, $\mathsmaller{T_{Q}^{X}T_{X}^{Q}}$, are below certain thresholds $\{\epsilon_{val}^{\theta},\epsilon_{val}^{\varepsilon}\}$. 

%
%
%
%
%

\begin{table*}[t!]
\begin{minipage}[h]{.67\textwidth}
\centering
\bgroup 
\newcommand{\sepwidth}{1pt} 
\newcommand{\shortsepwidth}{1pt} 
\newcommand{\vertsepwidth}{.5pt}
\setlength{\belowrulesep}{0pt} 
\setlength{\aboverulesep}{.25pt}
\def\arraystretch{1.15}
\fontsize{7.25pt}{8.5pt}\selectfont

\begin{tabular}{|c!{\vrule width \vertsepwidth}c|c|c|c|c|c|c|c|c|c!{\vrule width \vertsepwidth}}

\specialrule{\sepwidth}{0pt}{1pt}


{$Data$} & \multicolumn{7}{c|}{{$\mathbf{Appearance~Reliant}$ (RGB or RGB-D)}}& \multicolumn{3}{c|}{{$\mathbf{Depth-Only}$}}\tabularnewline
\cline{1-11}   
{$Approach$}  & \multicolumn{8}{c|}{{$\mathbf{Reconstruction~Truth~Needed~for~Relocalization}$}} & \multicolumn{2}{c|}{{$\mathbf{Retrieval}$}}\tabularnewline
\hline  


\multicolumn{1}{|c|}{{$Method$}} & {${Spr}$\cite{shotton2013score}} & {\cite{brachmann2016uncertainty}${C}$} & {${DSc}$\cite{brachmann2016dsac}} & {{\cite{shotton2013score}}} & {\cite{guzman2014MO}} & {{\cite{valentin2015exploiting}}} & {{\cite{brachmann2016uncertainty}}} & {\textit{D\cite{shotton2013score}}} & {${R}$} & {${VDR}$}\tabularnewline

\specialrule{\sepwidth}{0pt}{1pt}

{$Chess$}  & {70.7} & {94.9} & {97.4} & {92.6} & {96} & {99.4}& \textbf{99.6} & {82.7} & {97.3} & {99.5}\tabularnewline
\cline{1-11}   
  {$Fire$}  & {49.9} & {73.5} & {74.3} & {82.9} & {90} & {94.6}& {94.0} & {44.7} & {92.3} & \textbf{97.8}\tabularnewline 
\cline{1-11}
 {$Heads$}  & {67.6} & {48.1} & {71.7} & {49.4} & {56} & {95.9} & {89.3} & {27.0} & {93.5} & \textbf{98.9}\tabularnewline  
\cline{1-11}   
 {$Office$}  & {36.6} & {53.2} & {71.2} & {74.9} & {92} & {97.0} & {93.4} & {65.5} & {89.7} & \textbf{98.4}\tabularnewline 
\cline{1-11}   
 {$Pumpkin$}  & {21.3} & {54.5} & {53.6} & {73.7} & {80} & \textbf{85.1}& {77.6} & {15.1} & {78.3} & {82.8}\tabularnewline 
\cline{1-11}   
 {$Kitchen$}  & {29.8} & {42.2} & {51.2} & {71.8} & {86} & {89.3}& {91.1} & {61.3} & {87.9} & \textbf{93.7}\tabularnewline  
\cline{1-11}  
 {$Stairs$}  & {9.2} & {20.1} & {4.5} & {27.8} & {55} & {63.4}& \textbf{71.7} & {13.6}  & {54.8} & {61.0}\tabularnewline 
\hline
\hline
{$Average$} & 40.7 & 55.2 & 60.1 & 67.6 & 79.3 & 89.2 & 88.1 & 44.3 & 84.8 & \textbf{90.3} \\
\hline
\hline
{$Combine$} & {38.6}& {55.2} & {62.5} & {-} & {-} & {-}& {-} & {-} & {84.8} & \textbf{90.4} \tabularnewline
\specialrule{\sepwidth}{0pt}{1pt}
\end{tabular}
\egroup 

\caption{ \label{main_table}~The presented approaches (\emph{R} , \emph{VDR}) are compared with baselines through localization accuracies on the standard 7-scenes datasets from \cite{glocker2013real,shotton2013score}. All methods utilize RGB-D data during training, except \cite{shotton2013score} D, and our \emph{R} and  \emph{VDR}, which are based on range / depth data. During test time, the three leftmost approaches only take RGB images as input, while the three rightmost approaches only take range / depth images - the rest operate on RGB-D. \textit{Average} indicates the average among the $7$ datasets. \textit{Combine} indicates performance when jointly considering all $7$ scenes as a single database. VDR outperforms all the RGB-D approaches while using depth information \textit{only}. R performs very well as well, outperforming all but two RGB-D approaches. \vspace*{-1em}}
\end{minipage}
\hspace{.2em}
\begin{minipage}[h]{0.28\textwidth}
\centering
\includegraphics[width=1\textwidth]{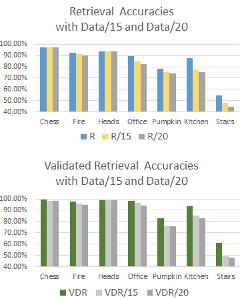}
\captionof{figure}{Accuracies with significantly sparser acquisition. Database sizes were reduced to $1/15$ and $1/20$.}
\label{fig:sparsedata}
\end{minipage}
\vspace{-1.5em}
\end{table*}

\subsection{Further details and discussion\label{subsec:details}}

 The approach is amenable to any boundary-preserving patch segmentation scheme, as long as it results in superpixels / patches that are geometrically regularized for smoothness and compactness. For example \cite{papon_VCCS2013}, which segments volumetrically, could be used while working with point clouds; and surface segmentation schemes such as one presented in \cite{GASP2014_3DV} could be employed when working with depth / range images. Both \cite{GASP2014_3DV} and \cite{papon_VCCS2013} performed well in our experiments. Starting with the base level segmentation, a segmentation hierarchy can be built in either fine to coarse (agglomerative), or coarse to fine (divisive) fashion. Each successive level has patches reduced (increased, in case of divisive) by a constant factor - this can be approximately ensured by employing K-Means in 3D with near uniform surface component seeding (any resultant patches below a certain size / surface area are merged back). We used four levels of segmentation hierarchy ($H=4$). The number of mixture components were also kept fixed, $G = 1250$. The GMMs were learnt through an expectation maximization scheme, and the mixture components were initialized from the result an iteration of K-Means++ procedure. Our empirical analysis indicated the learnt feature space distribution to be general for similar sensor types \footnote{~Sufficient number of GMM components should be utilized to span the extent of the geometric feature space. This is a function of maximum scene scale captured, and thus sensor range.}. Figure~\ref{fig:ablations} suggests that as well. In fact, a single set of Gaussian mixture (and ICA) models were utilized for all the experiments shown in the article (except Figure~\ref{fig:ablations}). 


 In practice, for efficiency, while encoding feature space statistics (Section \ref{subsec:encoding}), it suffices to approximately ascertain  $F_{h}^{X}$ by sampling patches from $X_h$, and subsequently sampling the neighborhoods of the sampled patches. This also partly corroborates our assertion that the methodology is robust to occlusions. \cmmnt{ The neighborhood sizes for all patches $x_{i}\in X$ were kept constant - $\mathcal{N}_{x_{i}}$ was determined simply by radius search about $l_{x_{i}}$.We used a fixed radius size ($3$ meters) in all experiments.} Databases were indexed as KD-trees. Our current straight up implementation is not optimized for efficiency (on a $4.2$ GHz, $4$ core setup, \ref{subsec:feat_representation} -  \ref{subsec:dpp} takes $\sim$ $.3$ ms, $1000$ superpixels),  though the methodology is GPU parallelizable. Most of the procedures outlined in Sections \ref{subsec:feat_representation}, \ref{subsec:encoding}, \ref{subsec:retrieval}, \ref{subsec:dpp} and  \ref{subsec:validation} can be GPU paralellized in a straightforward fashion. The computational bottleneck arises during validation, which is quadratic in number of superpixels ($\sim 1$s for segmentation with 1000 superpixels at finest level, but again naturally parallelizable). Note that it suffices to validate at a coarse hierarchical level ($\sim 250$ superpixels) --- the result, $T_{Q}^{X}$, can then be used as reliable initialization and be quickly refined iteratively as per task.

\section{Experiments\label{experiments}}
In all experiments, the method indicated \singlestrquotes{\emph{R}} refers to our retrieval approach (till Section \ref{subsec:encoding}), without the diverse subset selection and validation steps. \singlestrquotes{\emph{DR}} refers to our approach till Section \ref{subsec:dpp}, with diversification but without the validation step. \singlestrquotes{\emph{VDR}} would then refer to the complete approach, resulting in the set $\mathcal{{C}}_{vld}$ - diverse retrievals which have been validated through rigid overlap consistency. The retrievals in both the sets $\mathcal{{C}}$ and $\mathcal{{C}}_{vld}$ follow the same order (by $\mathbf{s}(X,Y)$) as they appear in the initial retrieval set $\mathcal{{R}}$. All analysis is done on the top few results from these sets.

The retrieval and association problems can be subjective - two views with only partially overlapping geometric content can be evaluated differently by users. We employed an objective measure - evaluating our retrieval approach on a sensor relocalization task. We utilized the 7-scenes datasets from \cite{glocker2013real,shotton2013score}, the standard benchmark for indoor RGB/RGB-D relocalization. The objective is to localize the sensor (ascertain pose) with respect to the workspace within the maximal allowable translation and orientation errors (5\,cm and 5~deg respectively). The datasets are collected from different workspaces (although some scenes in \textit{Redkitchen} and \textit{Pumpkin} are quite similar). Standard train - test splits are provided, with the viewpoints in the test set differing significantly from the training set. This makes it most appropriate for use in the evaluation \footnote{~As opposed to mapping, visual odometry or semantic scene datasets such as \cite{sturm2012RGBDbenchmark, handa2014benchmark, xiao2013sun3d, dai2017scannet, handa2015scenenet}. These either do not have enough loop closures and/or are synthetic, or lack ground truth for quantitative evaluation or standard train-test splits for loop closure.}. 7-scenes also provide additional training information - global sensor pose annotations, as well as reconstructed volumetric workspace models.

In our approach, \emph{R, DR and VDR}, depth images for training were simply encoded as an unordered view-signature database. A given query image from the test split was localized by computing the relative transform with respect to the top retrieval (in the sets $\mathcal{{R}}$, $\mathcal{{C}}$ and $\mathcal{{C}}_{vld}$ respectively), and the localization accuracy was computed by evaluating the disparity between the estimated and ground truth relative poses. Same as in the baselines, 5\,cm and 5~deg are the allowable error. Our approach did not require additional information accompanying the datasets to operate (pose truth annotations and workspace reconstructs). Importantly, it also did not require specific training for each dataset. This differs from most of our baselines which required some additional information or dataset-specific training.

\textbf{\emph{Baselines}:} We compare our approach against many baselines.  Approaches like \cite{shotton2013real, valentin2013mesh, guzman2014MO, kendall2017geometric, brachmann2016dsac, walch2017imagelocLSTM, melekhovYKR17} require additional information and dataset specific training. They rely on annotations, workspace models, and involve regression against absolute sensor poses or 3D coordinates of pixels. Deep-CNN based regressors have been proposed as well, such as \cite{kendall2017geometric, brachmann2016dsac, walch2017imagelocLSTM, melekhovYKR17}. Such approaches can overfit on the training data, and are difficult to generalize to scenes that are not similar to the training. Some baseline results were not shown in Table \ref{main_table} --- \cite{glocker2013real}, which presents a random ferns based retrieval method over RGB-D, report accuracies differently; but they indicate the achieved results to be weaker than some of the baselines considered in Table \ref{main_table}. Methods like \cite{kendall2017geometric, walch2017imagelocLSTM, melekhovYKR17} report localization accuracies as median errors - since the lowest reported median errors, that we are aware of, are greater than $10$ cm (translation, implicitly includes orientation errors as well), these methods are also not as accurate as some of the baselines in Table \ref{main_table}. Approaches \cite{li2015rgbdReloc} and \textit{Sparse} \cite{shotton2013score} employ frame to model matching for relocalization. They match local features from the query frame to a global feature cloud accumulated and reconstructed \textit{a priori} from the training data and the pose ground-truth annotations. \cite{li2015rgbdReloc} shows nice results, though we were unable to obtain exact numbers from the authors. However, \emph{VDR} in Table~\ref{main_table} does seem to perform better than \cite{li2015rgbdReloc} in 4 out of 7 datasets in comparison. \emph{VDR} also seems to outperform \cite{li2015rgbdReloc} in at least 6 out of 7 datasets when only 1/15 of the training data is used (Figure \ref{fig:sparsedata}). All the aforementioned approaches are appearance-reliant as well (except \cite{shotton2013score} which additionaly present a depth only variant). We also tried a retrieval methodology similar to ours with local 3D geometric point-features (such as \cite{tombari2010SHOT}), but their performance was worse than those shown in Table \ref{main_table}.

\begin{figure*}[t!]
\includegraphics[width=1\textwidth]{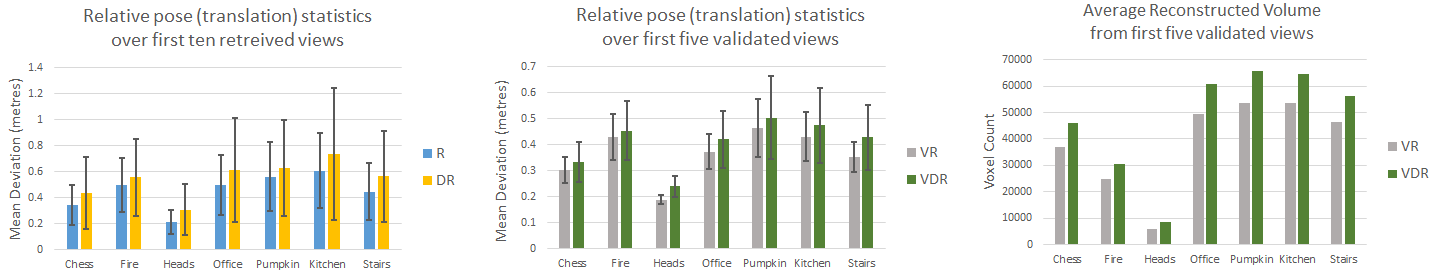}\\[-14pt] 
\protect\caption{\label{fig:diversity}~Quantifying diversity. Left, Middle: The average relative translation of the retrieved views with respect to the queried view. One can see DR improves diversity over R, and VDR improves over VR. Right: Efficacy of diverse viewpoints for reconstruction task. The average number of voxels (in a 8 $cm^3$ occupancy grid) occupied by ground truth reconstructs from the first five validated retrievals from VR and VDR are plotted. From the same number of initial views, VDR results in richer reconstructs that capture significantly more voxels in the scene.}
\vspace*{-1em}
\end{figure*}

%

As Table \ref{main_table} shows, \emph{VDR} achieved state-of-the-art results through pure geometry alone - without needing any additional annotations, assumptions or appearance features. Equally promising were the results from \emph{R} which were obtained by simply using the first retrieval in $\mathcal{R}$ (no diversification or validation), which were better than all baselines but two. \emph{DR} gives the same results as \emph{R} in the relocalization experiments and is hence not shown. This is because the accuracies were evaluated with respect to only the top retrieval - this is the same for \emph{R} and \emph{DR} since the greedy algorithm we used for k-DPP automatically selects the top-scoring retrieval as the first one. These results support our hypothesis that macro-level 3D geometry holds immense discriminative information. 

In the last row of Table \ref{main_table}, we combined all training data from the $7$ datasets into one single database, and evaluated accuracies of the combined test splits. As can be seen, the results held up quite well in the combined experiment, when the database size and complexity (variety, aliasing) was drastically increased. 

We also tabulated the affect of significantly reducing the database sizes - by re-evaluating results with databases built from only $1/15^{th}$ and $1/20^{th}$ of the available train-splits for each dataset. With a much sparser coverage of the environment, both retrieval and subsequent validation and localization becomes much more difficult. The frames were sampled at uniform intervals, thus may have steep viewpoint changes, much reduced content overlap and significantly increased occlusions. As Figure \ref{fig:sparsedata} indicates, the accuracies of both \emph{R} and \emph{VDR} held up quite well. This is indicative of the approach's robustness to these practical challenges.


The approach generalizes well. Our experiments do not suggest a need for scene specific training - a single set of learnt gaussian mixtures and ICA projection matrices were utilized in all our experiments (except Figure \ref{fig:ablations}). The training data was taken from the train split of Redkitchen in \cite{shotton2013score}, and from datasets in \cite{xiao2013sun3d, sturm2012RGBDbenchmark}, a reasonably rich and diverse set of samples. Figure \ref{fig:ablations} shows the robustness of the GMM parameters with respect to the dataset used to train it. As can be seen the results stay consistent.

\begin{figure*}[t!]
\includegraphics[width=1\textwidth]{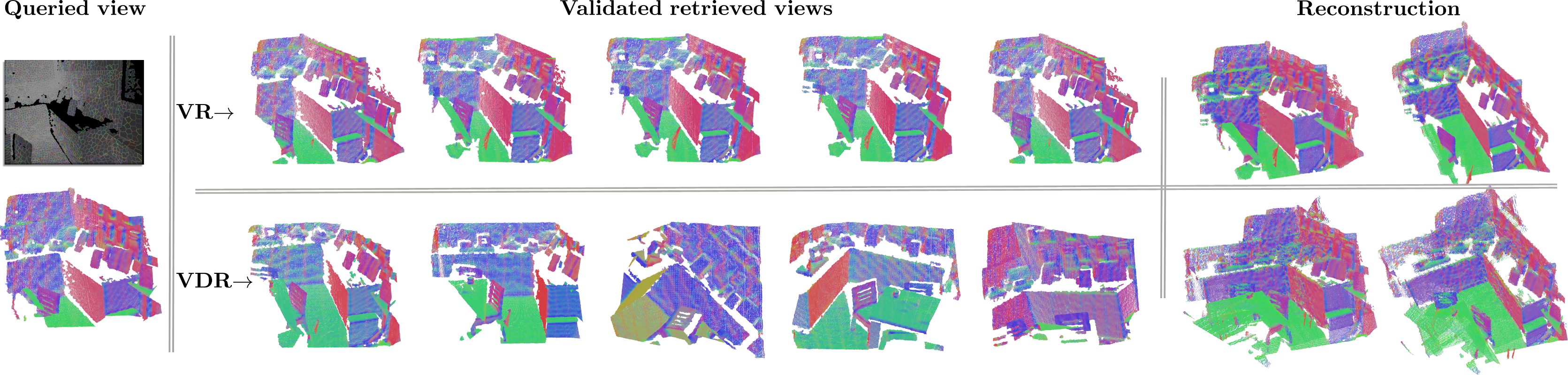}
\hrule
\includegraphics[width=1\textwidth]{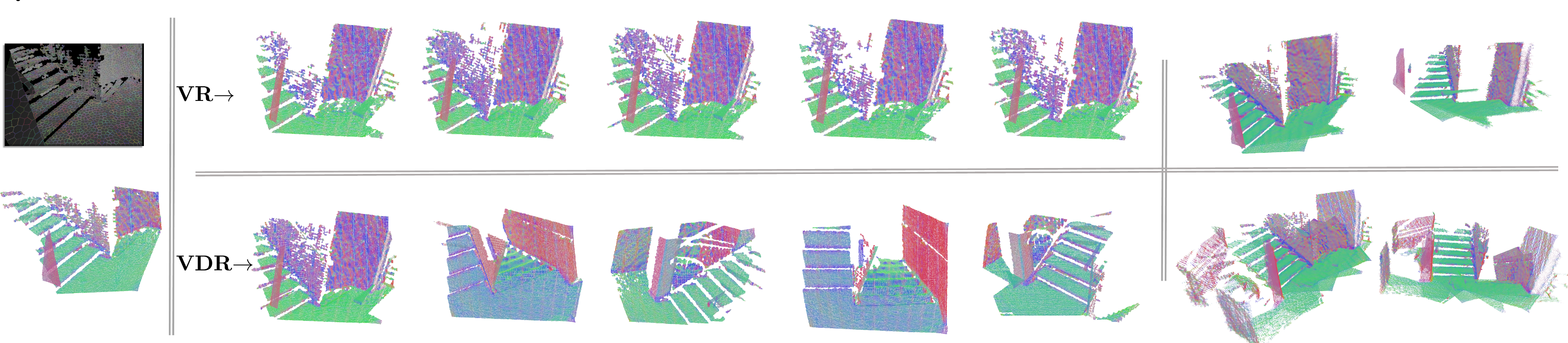}\\[-18pt]
\protect\caption{\label{fig:retrieval_example} Retrievals and reconstructions are shown for two example scenes (top and bottom). For each scene, queried view is shown on the left as a depth image with overlaid patch boundaries. Views on top row are the top-five retrieved and validated views without using DPP. Views on the bottom row are the top-five validated views with DPP. Reconstructed scene models from the respective sets are shown on the right from two perspectives. Note that viewpoints vary significantly in the diversified retrievals, and results in a much larger reconstructed volume (over 1.5x).\vspace*{-1em}}
\end{figure*}

Finally, we conducted experiments to quantify the effect of our diversification approach, and its role in generating significantly richer reconstructions. As Figure \ref{fig:diversity}(left, middle) show, the diversity of retrieved viewpoints is greatly improved due to our DPP-based diversification. Note that DR and VDR select views which are not only further off than the queries (higher relative mean), but result in view sets which have significantly more viewpoint variance amongst themselves as well (significantly higher standard deviations). And as Figure \ref{fig:diversity}(right) shows, the reconstruction volume improves significantly when a diversified set of views is utilized. Figure~\ref{fig:retrieval_example} shows a qualitative example. One can see that the diversified retrievals are significantly more diverse, from varied viewpoints, and are resulting in an appreciably richer reconstruction. In general, retrieval as well as geometric diversity is often desirable - apart from reconstruction, it would prove useful in other tasks such as structure and semantic analysis, and \singlestrquotes{human in loop} selection tasks.

\section{Conclusion \label{sec:conclusion}}
We presented a robust solution to the problems of measuring 3D geometric similarity between 3D range images or point clouds, and determining whether they come from the same scene. A general-purpose retrieval approach was proposed, based on encoding (FV) of viewpoint-invariant features that are hand-crafted to capture 3D geometry at macro scales. The approach performed well in real world settings - including ones that involved sharp viewpoint changes, partially overlapping and occluded content. It scaled well, and did not require scene-specific training - making it useful in a variety of scenarios. As experiments established, the approach is powerful and did better than specifically fitted solutions such as CNNs trained on RGB or RGB-D data. Furthermore, we introduced a way to obtain \emph{geometrically diverse} retrievals (DPP), and showed how such retrievals can help generate richer reconstructions. Interestingly, in contrast to CNN approaches which begin with a local neighborhood, our approach utilized macro scale features from start. The combination of both paradigms would be explored in future work, for this and other tasks involving 3D recognition.
\begingroup
\vspace{-.15em}
\subsection*{Acknowledgements}
RS and FL are supported in part by NSF grant IIS-1464371.
\endgroup

\bibliographystyle{abbrv_named}

{
\footnotesize{\bibliography{rbib.bib}}
}
\end{document}